\documentclass{article}  

\usepackage{geometry}
\usepackage{url}

\usepackage{graphics} 
\usepackage[pdftex]{graphicx}

\usepackage{amsmath} 
\usepackage{amssymb}  
\usepackage{gensymb}
\usepackage{array}
\usepackage{multirow}

\title{Design and Characterization of 3D Printed, Open-Source Actuators for Legged Locomotion
}

\author{Karthik~Urs$^{1}$,
        Challen~Enninful~Adu$^{1}$,
        Elliott~J.~Rouse$^{1,2}$,
        and~Talia~Y.~Moore$^{1,2,3}$%
\thanks{$^{1}$ K. Urs, C. Enninful Adu, E. J. Rouse, and T. Y. Moore are with the Robotics Institute,
University of Michigan, Ann Arbor, MI 48109 USA email: \{ursk, enninful, ejrouse, taliaym\} @umich.edu.}
\thanks{E. Rouse and T.Y. Moore are with Mechanical Engineering, Robotics,
University of Michigan, Ann Arbor, MI 48109 USA.}
\thanks{T.Y. Moore is affiliated with Ecology and Evolutionary Biology, and the Museum of Zoology, University of Michigan, Ann Arbor, MI 48109 USA.}
}

\begin{document}

\maketitle

\begin{abstract}

Impressive animal locomotion capabilities are mediated by the co-evolution of the skeletal morphology and muscular properties.
Legged robot performance would also likely benefit from the co-optimization of actuators and leg morphology.
However, development of custom actuators for legged robots is often expensive and time consuming, which discourages roboticists from pursuing performance gains afforded by application-specific actuator optimization.
This paper presents open-source designs for two quasi-direct-drive actuators with performance regimes appropriate for an 8--15~kg robot, built completely with off the shelf and 3D-printed components for less than \$200 USD each.
The mechanical, electrical, and thermal properties of each actuator are characterized and compared to benchmark data.
Actuators subjected to 420k strides of gait data experienced only a 2\% reduction in efficiency and 26~mrad in backlash growth, demonstrating viability for rigorous and sustained research applications.
We present a thermal solution that nearly doubles the thermally-driven torque limits of our plastic actuator design.
The performance results are comparable to traditional metallic actuators for use in high-speed legged robots of the same scale.
These 3D printed designs demonstrate an approach for designing and characterizing low-cost, highly customizable, and highly reproducible actuators, democratizing the field of actuator design and enabling co-design and optimization of actuators and robot legs.
\end{abstract}


\section{INTRODUCTION}
\label{sec:intro}

{Bioinspired} legged robots have shown impressive agility, terrain traversal ability, and robustness \cite{minicheetah, ANYmal, SubT, azocar2020design}.
However, their performance capabilities are limited when compared to the diversity of terrestrial animals and their many forms of legged locomotion. 
The morphological features of robots also differ greatly from their animal inspirations.
The most popular mechanical designs have maximized simplicity and generality, typically featuring identical actuators and two-link legs across the robot \cite{raw_patel2019}.

In animals, locomotor performance (\emph{i.e.}, maximum speed, acceleration, cost of transport, and maneuverability) is largely determined by the structure of the musculoskeletal system, which varies greatly across species \cite{Biewener,polly_limbs_2007}.
Analogously, morphological factors in quadrupedal robot design, such as leg shape and orientation, have been optimized to increase whole robot performance \cite{de_vincenti_control-aware_2021, raw_patel2019, ha_task-based_2016}.
Separately, actuator design has been optimized, but many still maintain generality for use in multiple leg joints \cite{Ding2017}.

In animals, selective pressures have favored the co-evolution of skeletal morphology and muscular properties to enable performance capabilities key to animal survival and success \cite{Biewener}.
Similarly, robot performance would likely benefit from coordinated optimization of leg and actuator design parameters (\textit{e.g.} motor selection, transmission ratio) \cite{chadwick2020vitruvio}.
However, traditionally-manufactured actuators for legged robots are often expensive and time-consuming to develop, usually requiring multi-axis CNC machining of custom parts. 
Despite clear performance benefits demonstrated in robots optimized for specific tasks (e.g., manufacturing, drones), upfront costs currently discourage roboticists from pursuing the performance gains afforded by co-designing application-specific robotic legs and actuators.

\begin{figure}[t]
\begin{center}
    \includegraphics[width=1\columnwidth]{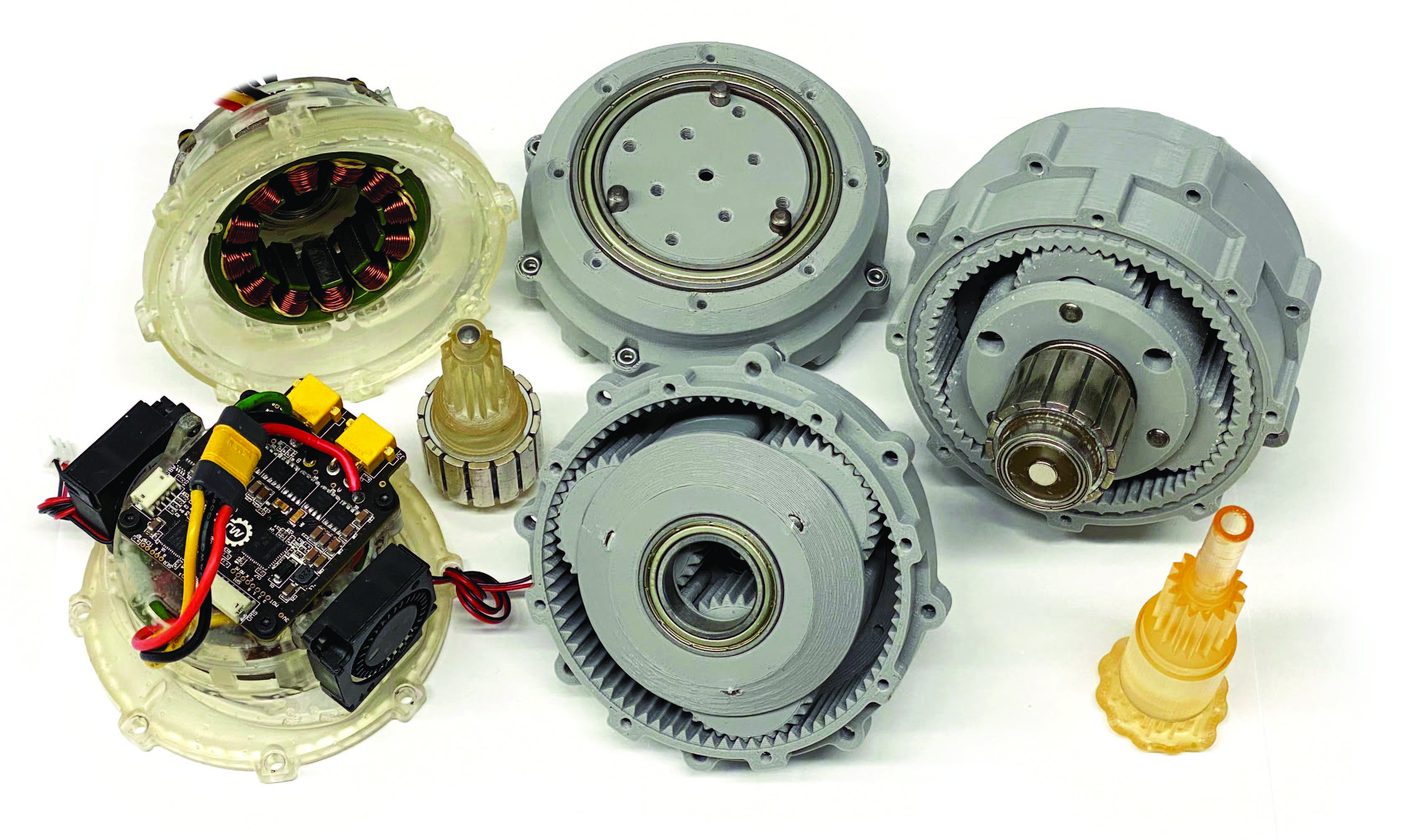}
    \caption{Actuator Hardware. Amber components are SLA-HT and grey components are FDM-PLA printed. A full actuator can be printed in 14 hours. A full actuator consists of one Input Assembly (primarily amber, left) and one Transmission Assembly (primarily grey, right).}
    \label{fig:photo}
\end{center}
\end{figure}

Rapid prototyping helps reduce the cost and time for iterative and customized actuator design.
Researchers and hobbyists have leveraged 3D printing to design general-purpose actuators and transmissions \cite{novakova-marcincinova_intelligent_2011, noauthor_highgearratio_3DP_actuator,noauthor_hightorque_3DP_actuator}, including quasi-direct drive actuators \cite{noauthor_opentorque_nodate}. 
To the best of our knowledge, no 3D printed actuators have been characterized to rigorously understand their performance for legged robots.
Additionally, the current designs do not quantitatively address the thermal issues or lifecycle limits that could arise from their plastic construction.
Other work has more rigorously addressed lifecycle limits for spur gears, but is limited in scope to steady-state rotation of a single gear pair \cite{zhang_physical_2020}.
\begin{figure*}[t]
    \centering
    \includegraphics[width=\textwidth]{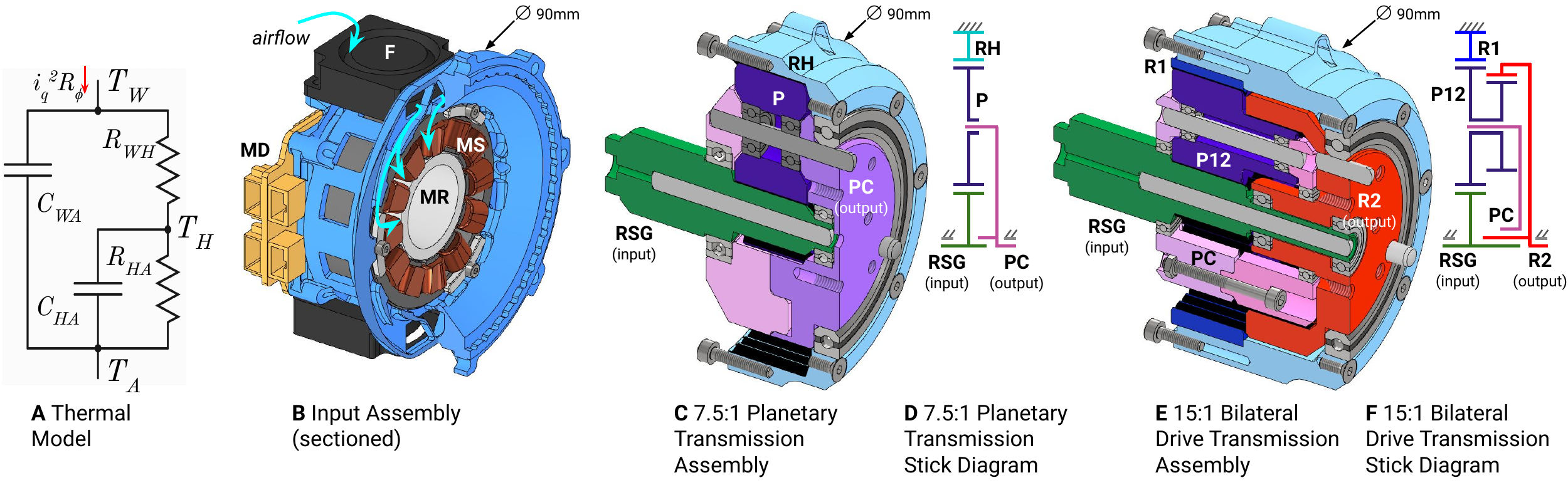}
    \caption{\textbf{A} Resistive-capacitive thermal model of motor in actuator. \textbf{B} Input Assembly with airflow from fans ($F$) highlight and motor driver ($MD$) mounted; 
    motor stator ($MS$) and motor rotor ($MR$, floating) within high-temperature stereolithography printed housing (blue). 
    \textbf{C} 7.5:1 Planetary Transmission Assembly and \textbf{D} stick diagram. \textbf{E} 15:1 Bilateral Drive Transmission and \textbf{F} stick diagrams.
    Transmission symbols: Rotor-sun gear ($RSG$), planet carrier ($PC$), planet ($P$), ring gear housing ($RH$), planet level 1 \& 2 ($P12$), ring level 1 ($R1$), ring level 2 ($R2$).}
    \label{fig:mech_design}
\end{figure*}

To address these challenges and facilitate rapid and inexpensive actuator customization for optimized designs, we present two fully-characterized, open-source quasi-direct-drive actuators made with only off-the-shelf components and 3D printed parts.
These actuators can be made with only \$200 US in materials, commonplace 3D printers, and basic tools (Fig. \ref{fig:photo}). 
High torque output while maintaining backdrivability and low inertia potentially make these actuators well-suited for agile and robust legged robots \cite{impactmitigation, Urs2022Framework}.
The ease of reproduction and customization also make them useful for building legged robot models to study animal locomotion \cite{long2012darwin, nyakatura_reverse-engineering_2019}.
Our characterization demonstrates that 3D printed actuators can be used directly or as fully-functioning prototypes for novel robot platforms.

We present the following key contributions:

\begin{enumerate}

\item We demonstrate the viability of 3D printed actuators for legged robotics through rigorous characterization (Section \ref{sec:charmethod}).
Empirically determined mechanical, electrical, and thermal properties are presented and compared to benchmark data (Section \ref{sec:charresults});

\item We present a thermal solution for plastic actuator parts (Section \ref{subsubsec:thermaldes}) that raises thermally-driven performance limits to near-parity with metallic designs (Section \ref{subsubsec:resultthermdyn});

\item We demonstrate the durability of 3D printed actuators by measuring changes in performance over 420,000 strides of simulated gait execution (Section \ref{subsec:resultlifecycle});

\item We facilitate the rapid development of new 3D printed actuators by making the mechanical, electrical, and software designs completely free and open source. 


\end{enumerate}


\section{Actuator Design}
\label{sec:design}


Our actuators are designed for an 8--15~kg quadrupedal robot --- a useful scale for indoor testing, analogous to a small goat or dog.
At this scale, electromagnetic actuators, particularly 
Permanent Magnet Synchronous motors (PMSMs), lend themselves to use in legged robots due to their availability in a variety of sizes, ease of control, and useful performance characteristics \cite{minitaur}. 
While Series Elastic Actuators (SEAs) have been used effectively to mitigate impact forces, improve torque fidelity, and store energy \cite{ANYmal, pratt1997stiffness, azocar2020design, ruppert2019series}, Quasi-Direct-Drive (QDD) actuators have been particularly effective at high speed, responsive locomotion \cite{minicheetah, Ding2017, pennjerboa}. 
QDD actuators can also virtually mimic the elasticity of SEAs via impedance control or other techniques \cite{Kalouche_GOAT} and achieve impact mitigation via lower output mechanical impedance (OMI) at frequencies relevant to animal locomotion \cite{impactmitigation, zhu2021design}.


We seek to follow the QDD actuator design philosophy, but increase reproducibility and design flexibility by exploiting the use of 3D printing.
Based on existing robots at the 8--15~kg scale, actuators should be capable of 5--15~Nm of torque and speeds of 15--35~rad/s with output inertia in the $10^{-4}$--$10^{-2}$~kgm$^2$ magnitude range \cite{minicheetah}.

\subsection{Motor and Transmission Co-Selection}
\label{subsec:coselection}

For legged robots that perform highly dynamic motions, it is important that their actuators feature high torque density and power, as well as increased responsiveness and impact mitigation ability \cite{impactmitigation}. Thus motors with a high torque constant $K_T$, high motor constant $K_M$, low inertia $J_m$, and low mass $m$ are desired. To select a motor with the right balance of these properties and the appropriate transmission to complement the motor, we used the design methodology detailed in \cite{Urs2022Framework}. 
To ensure compatibility with 3D printing manufacturing, only direct, planetary, and bilateral drive  \cite{Matsuki2019} transmissions with ratios between 1:1 and 30:1 were considered for their exclusive use of rigid elements.

\begin{table}[h]
    \caption{T-Motor RI50 motor properties, summarized from \cite{Urs2022Framework}.
    }
    \centering
    
    \begin{tabular}{rcl|ccc}
     & $  $  &  & RI50 \\ \hline \hline
Winding Style & $  $  &  & Wye \\
Torque Constant & $ K_T $  & [Nm/A] & 0.105 \\
Motor Constant & $ K_M $  & [Nm/W$^{0.5}$] & 0.12 \\
Motor Inertia & $ J_m $  & [kgm$^2$] & $9.01 \times 10^{-6}$ \\
Mass & $ m $  & [g] & 193 \\
Responsiveness Metric  \cite{Urs2022Framework} & $ S_M $  & [ms] & 0.67 \\ 
Torque-Specific Inertia \cite{Urs2022Framework} & $ S_T $  & [g(A/N)$^2$] & 0.82 \\ \hline 
    \label{tab:motordata}
    \end{tabular}
    
\end{table}

The interior-rotating T-Motor RI50 is a motor that has a more useful balance between torque constant, motor constant, and inertia 
than a dimensional scaling analysis would suggest for its small gap-radius \cite{impactmitigation, Urs2022Framework}, and is available for \$62. 
The RI50's lower $K_T$ and $K_M$ can be compensated for by a higher gear ratio
, while still maintaining a lower reflected output inertia 
compared to an actuator with the U8 or equivalent motor \cite{Urs2022Framework}.
This advantage was clearly indicated by the $S_M$ and $S_T$ selection metrics and is likely due to a higher coil turn count and an inner-rotating geometric advantage for the stator \cite{Urs2022Framework}.
Leveraging this advantage, the RI50 motor was selected and paired with two different transmission styles: planetary 7.5:1, and bilateral drive 15.04:1 (abbr. 15:1), similar to the configuration in \cite{ding_design_2017}. 
These ratios yield performance regimes appropriate for an 8--15~kg quadrupedal robot based on benchmarks (Table \ref{tab:actuator_comparison}).




\subsection{Mechanical Design}
\label{subsubsec:mechdes}

All non-off-the-shelf components in the design are 3D printed. 
The majority of the parts are fused-deposition (FDM) printed with polylactic acid (PLA) plastic (Hatchbox, Pomona, CA, USA) on a Prusa MK3S+ printer (Prague, Czech Republic).
These PLA parts use approximately \$3.50 in material.
Though FDM-PLA parts lack mechanical strength compared to molded plastic or metals \cite{ngo2018additive}, designs that avoid stress concentrations and make best use of print orientation have proved to be sufficiently strong for the application.
Because temperatures near the motor can exceed the glass transition temperature of PLA, some parts (denoted SLA-HT below) are printed via stereolithography (SLA) in High Temp Resin with a Formlabs Form 3 printer (Formlabs, Somerville, MA). These SLA-HT parts use $\sim$\$16 in material. The actuator is split into two subassemblies (Fig. \ref{fig:mech_design}). 
The Input Assembly consists of the RI50 motor stator, cooling solution (see Sections \ref{subsubsec:thermaldes} and \ref{subsec:resultthermal}), and open-source motor driver (\$84, moteus r4.5, mjbots, Cambridge, MA, USA).  All these components are mounted to a SLA-HT motor housing.

The RI50 motor rotor is attached to the SLA-HT rotor-sun gear (RSG), which is reinforced with a steel dowel pin and carries a magnet to interact with an encoder, and also bridges the Input Assembly and Transmission Assembly.
All components in the Transmission Assembly apart from bearings, pins, and fasteners are FDM printed in PLA.
In the instance of the 7.5:1 planetary transmission, the planet carrier is itself the output.
For the bilateral drive transmission, the output is the secondary ring gear.
All gears use a $30\degree$ pressure angle involute profile to increase strength \cite{rick_miller_designing_2017}.
Bearings amount to $\sim$\$15.
The two main housings register with each other on an annular boss and are axially mated together with eight screws.
Fasteners and pins amount to $\sim$\$8.

The bilateral drive is a compound-planetary transmission that follows the topology commonly known as Wolfrom \cite{wolfrom_wirkungsgrad_1912}.
These transmissions can achieve very high ratios in compact packaging, but typically have low efficiency compared to multi-stage planetary transmissions of the same ratio \cite{garcia_compact_2020}. 
To overcome this challenge, we have implemented an established efficiency optimization method with some additional constraints on allowable ranges for parameters to generate gear geometry for the 15:1 transmission \cite{Matsuki2019}.



The design of functional and reliable 3D printed components required some special considerations. 
FDM printed parts are strongest in the print-layer plane (\textit{i.e.}, XY plane) \cite{santana_study_2017, gorski_influence_2013}, so gears were printed with their axis parallel to the Z axis---tangential forces from torque transmission then stays in the strong plane.
Components for which not all loads can be contained in the strong plane, or where shear forces are present in the strong plane, were then reinforced by incorporating metal pins to avoid layer delamination.
Delamination due to tension was avoided by use of compression preload---this preload is generated with screws between the two halves of the planet carrier.
Whenever possible, we include a pocket to embed a metallic nut with threads for bolts and screws.
Heat-set inserts are another solution for threaded holes, but they cannot be placed near the edge of parts without avoiding warping during the heat-setting process in our experience.


Geometric mating features help distribute stress and aid in assembly and disassembly.
For example, splines help prevent rotation between pieces when assembled --- the R1 element of the 15:1 transmission is torsionally retained this way (Fig. \ref{fig:mech_design}).
The two planet carrier halves have pin and socket features to ensure proper alignment (\textit{i.e}., keeping all rotational axes parallel).
Ejection ports where pins may be strategically pushed through also aided in disassembling press-fit or close-fit features.

\subsection{Thermal Design}
\label{subsubsec:thermaldes}

Quasi-direct-drive actuators require low gear-ratios and therefore low torque amplification.
Maximizing torque density then requires driving the motors with high currents, producing significant joule heating.
Thermal management is critical to allow useful continuous actuator usage and high-power peak conditions, especially with plastic components
\cite{ngo2018additive, flaata2017thermal}.
Prototypes with passive cooling through plastic or adhered aluminum heatsinks proved to be insufficient to meet performance goals, so we designed an active cooling solution.

Legged robot actuators often need to exert torque without rotating (\textit{e.g.}, standing while carrying load or self-weight), so using the main actuator motor to move fans is insufficient; dedicated fans are necessary.
Two auxiliary centrifugal fans (\$5 ea, 30$\times$10mm) are mounted to the side of each actuator.
The output airflow from the fans is routed into the motor via ducting cavities in the housing (Fig. \ref{fig:mech_design}) and exhausted through ports on the other side of the motor, forcing the flow over the motor windings and rotor.


\section{Characterization Methods}
\label{sec:charmethod}


To empirically determine the mechanical, electrical, and thermal properties of the actuator useful for whole-robot design and modelling, we constructed a dynamometer to conduct characterization experiments (described in \cite{Urs2022Framework}).

\subsection{Motor Properties}
\label{subsec:methodpassive}
Properties of the RI50 motor were empirically determined in \cite{Urs2022Framework}. 
The methods are repeated here in brief for convenience:
\begin{itemize}
\item \textbf{Winding style:} A ``line-to-line'' current was applied through the motor to heat it while the spatial distribution was monitored with a thermal camera.
\item \textbf{Resistance ($R_\phi$):} Precision multimeter line-to-line measurement, converted to phase-frame.
\item \textbf{Inductance ($L_e$):} Precision LCR meter line-to-line measurement, converted to phase-frame.
\item \textbf{Torque Constant ($K_T$):} Various q-axis currents were commanded to the motor while it was held under stall. Actual q-axis current $i_q$ was measured by the motor driver along with torque by a transducer; a line was then fit to this data where the slope is $K_T$.
\item \textbf{Back EMF Constant ($K_B$):} The motor was back-driven by a cordless drill while the phase line-to-line voltages were measured with an oscilloscope.
\item \textbf{Motor Constant ($K_M$):} $K_M$ can then simply be calculated as $K_M= \sqrt{K_T K_B/R_{\phi}}$
\end{itemize}

\subsection{Dynamic Properties} 
\label{subsec:methoddynamic}

While the motor constants describe steady-state behavior, dynamic properties must be determined to aid in system modeling and control design.
In legged robots with very light leg links, actuator dynamics may dominate leg system dynamics \cite{impactmitigation}.


We determined the actuator dynamic properties via random-input testing to observe the frequency response of the system using a procedure described in detail in \cite{Urs2022Framework}.
In brief, an electrically disconnected actuator was mechanically driven with a low-pass-filtered torque signal produced by the driving actuator. 
A parametric transfer function estimate of the system was then fit to the data with a two-stage optimization procedure.

The experiment was repeated without the actuator so that dynamics of the dynamometer hardware (\textit{e.g.} shaft couplers) could be subtracted away.

\subsection{Thermal Modeling}
\label{subsec:methodthermal}
The temperature rise of the stator windings, which must be kept below 100 $^\circ$C to avoid premature winding thermal failure, is typically the limiting factor for actuator torque.
It is particularly critical to manage temperature rise in the primarily-plastic design, in which critical components may fail by warping due to heat.
To understand performance limitations and the benefit of the cooling solution (Section \ref{subsubsec:thermaldes}), we empirically identified a thermal model of the system.

\subsubsection{Electrical-Thermal Calibration and Measurement}
\label{subsubsec:methodthermistor}

To take temperature readings of a PMSM, thermal imaging followed by averaging over the winding visible regions is a reliable method to encapsulate spatial temperature distribution effects, unlike single-point measurements \cite{lee_empirical}.
However, the motor enclosure geometry determines air flow and blocks imaging temperature measurement.
Instead, we modeled the effect temperature has on winding electrical resistance, effectively using the motor as a thermistor.


To calibrate the ``thermistor-effect'' behavior of the motor, the windings were exposed to a thermal camera while a voltage was applied across two of the motor phases.
An in-line power meter measured the voltage $V$ across and current $i_{bus}$ through the phases, yielding the line-to-line winding electrical resistance $R_{ll} = V/i_{bus}$.
The average temperature of all the winding pixels (a roughly annular region in Figure \ref{fig:thermal_bw}) was plotted with 50-sample averaged resistances to yield a linear fit calibration.

\subsubsection{Thermal Dynamics}
\label{subsubsec:methodthermdyn}

We conducted a test similar to Section \ref{subsec:methoddynamic} to identify the thermal system.
Random voltages were programmatically generated and commanded to an adjustable power supply (1687B, B\&K Precision, Yorba Linda, CA, USA), which applied the voltage across two motor phases to heat the motor.
The random voltages were held for 20~s before changing.
Temperature was obtained via $R_{ll}$ and the thermistor-effect calibration.
To understand the relative improvement from the fans, this experiment was conducted with and without the fans powered for one hour from a cold start.

For identification purposes, we considered the system to be the following second-order ODE corresponding to the thermal circuit in Fig. \ref{fig:mech_design} A.
The input signal was the electrical power input $P = Vi_{bus}$ (which is all translated to heat), and the output signal was the temperature rise above ambient ($T_A = 25 \degree$C) as measured via the thermistor effect:
\begin{equation} \label{eq:thermal_tf}
    \frac{T_W (s) - T_A}{P(s)}= \frac{R_{HA} + R_{WH}+ C_{HA}R_{HA}R_{WH}s}{(C_{HA}R_{HA}s + 1)(C_{WA}R_{WH}s + 1)}
\end{equation}
where the $R$ and $C$ terms are defined in Figure \ref{fig:mech_design}.


\subsection{Power Efficiency}
\label{subsec:methodpower}

To characterize power efficiencies for our actuators, we conducted an experiment with a loading actuator under velocity control and a driving actuator under torque control.
A set of 1838 conditions for the 7.5:1 actuator and 1568 for the 15:1 actuator across positive and negative power regimes were performed. 
Electrical power meters collected voltage and current data on the DC bus, while the torque sensor and motor driver encoders collected mechanical power data.
To manage electrical transients, the loading actuator was ramped up to speed over 0.1~s, followed by a 0.1~s delay before the driving actuator was ramped up to torque over 0.1~s. 
The condition was held for 0.8~s before ramping down in reverse order.

\subsection{Lifecycle}
\label{subsec:methodlifecycle}

Lifecycle testing of 3D-printed gears has been conducted in \cite{zhang_physical_2020}, but we employ a test that more accurately captures the transient nature of legged locomotion.
We subjected two 7.5:1 actuators to a 57 hour lifecycle test in application-like loading.
We used 106 minutes of MIT Mini-Cheetah torque and velocity data from its 12 actuators while it traversed level ground in various gaits (standing, walking, trotting).
These data were ``played back'' on the dynamometer on loop: torque data were commanded to the driving actuator (tracking through current control) while the velocity data were commanded to the loading actuator (velocity controlled).
Every 41 minutes, the playback was interrupted with interval tests to gauge wear:

\begin{enumerate}
\item Power Efficiency: The same procedure described in Section \ref{subsec:methodpower} was repeated with a total of 24 torque-velocity conditions. 
The median of the torque estimates (i.e., $i_q K_{Ta}$) from the two actuators was used in place of the torque sensor, which was uninstalled for the test.
\item Steady-State Damping: While the loading actuator was placed in an idling state (motor phases open), the driving actuator was commanded with a series of velocities. 
The ratio of motor driver torque estimate to velocity yielded an effective damping coefficient.
\end{enumerate}


\subsection{Control Performance}
\label{subsec:methodcontrol}

PMSMs driven with FOC can use high-bandwidth current control as a proxy for torque control to enable responsive action \cite{mevey2009sensorless}.
To characterize the current controller response, we conducted a step-response test at input magnitudes of $i_q=[1,2,3,4,5,6]$~A.
Each condition was tested for 50 trials and sampled at 2.7~kHz to generate an averaged response profile.

We also conducted a random-input command test in the same manner described for the current controller (Section \ref{subsec:methoddynamic}), but with $i_q$, commands at amplitudes of $[2.0, 4.0, 6.0]$~A (low-pass filtered to 300~Hz) with command and sampling at 2.7~kHz.
The output signal was the motor driver's current measurement.
The random-input test was repeated on the 7.5:1 actuator to characterize the position controller.
Here, actuator position commands, $\theta_a$, were the input at amplitudes of $[\pi/8, \pi/4, \pi/2, \pi]$~rad (low-pass filtered to 60~Hz) with command and sampling at 1.1~kHz.
The output signal was the motor driver angle measurement.
For both of these random-input controller tests, results were analyzed in the frequency domain via the Matlab \texttt{spa()} method in order to find the controller half-power bandwidths.
\section{Characterization Results}
\label{sec:charresults}

Our characterization results show that the 3D printed actuators possess a favorable set of mechanical, electrical, and thermal properties despite the all-plastic, low-cost construction (Table \ref{tab:actuator_comparison}).
We present our data in comparison to a benchmark, the MIT Mini-Cheetah actuator, which is more compact, has a higher maximum speed, is deployed on a successful legged robot platform \cite{minicheetah}, and has also been adapted to other applications \cite{zhu2021design}. 
Although 2\% of the Mini-Cheetah velocity data used for lifecycle testing (Section \ref{subsec:methodlifecycle}, \ref{subsec:resultlifecycle}) exceeded the top speed of our actuator, other legged robots perform well in this lower speed regime \cite{ANYmal}.

Lifecycle testing on the 7.5:1 actuator revealed a key finding: the actuators are capable of demanding usage for a sustained period of time; the tested actuators did not experience any failures over the course of the 57 hour test. 
All intervals are reported to 95\% confidence ($2\sigma$) unless otherwise noted.

\begin{table}[]
    \setlength\tabcolsep{1.4pt}
    \begin{center}
        \caption{Key actuator performance metrics. Peak torques are thermally driven; maximum tested torques are lower. Note: Mini Cheetah (``MC'') peak torque duration was not provided}
        \begin{tabular}{r c l | c c c}
 & $ $ &  & 7.5 & 15 & MC \cite{minicheetah} \\  \hline \hline
Transmission Ratio & $ $ &  & 7.5:1 & 15.04:1 & 6:1 \\ 
Transmission Style & $ $ &  & Planetary & Wolfrom & Planetary \\ 
Eff. Torque Const. & $K_{Ta}$ & [Nm/A] & 0.79 & 1.58 & 0.84 \\ 
Eff. Motor Const. & $K_{Ma}$ & [Nm/W$^{\frac{1}{2}}$] & 0.88 & 1.76 & 1.38 \\ 
Actuator Inertia & $J_a$ & [gm$^2$] & $0.64$ & $2.2$ & $2.3$ \\ 
Mass & $m$ & [g] & 482 & 536 & 480 \\ 
Diameter $\times$ Length & & [mm] & $90\times70$ & $90\times90$ & $96\times40$ \\ 
Peak Torque (20s) & $ $ & [Nm] & 7.5 & 15 & \multirow{2}{*}{17} \\ 
Peak Torque (2s) & $ $ & [Nm] & 19.1 & 38.2 &  \\ 
Continuous Torque & $ $ & [Nm] & 4.4 & 8.8 & 6.9 \\ 
Max Speed & $ $ & [rad/s] & 21 & 11 & 40 \\ 
Thermal Resistance & $R_{th}$ & [$^\circ$C/W] & \multicolumn{2}{c}{2.27} & 1.23 \\ \hline
        \label{tab:actuator_comparison}
        \end{tabular}
    \end{center}
    
\end{table}

\subsection{Motor Properties}
\label{subsec:resultpassive}
The results from \cite{Urs2022Framework} are summarized here for convenience:

\begin{itemize}
\item \textbf{Winding style:} Wye.
\item \textbf{Resistance ($R_\phi$):} $R_\phi = 705\pm 3$ m$\Omega$
\item \textbf{Inductance ($L_e$):} $L_e = 2559\pm 2$ {$\mu$}H
\item \textbf{Torque Constant ($K_T$):} $K_T = 0.105\pm 0.002$ Nm/A
\item \textbf{Back EMF Constant ($K_B$):} $K_B = 0.094\pm 0.002$ Vs/rad
\item \textbf{Motor Constant ($K_M$):} $K_M = 0.118 \pm 0.003$ Nm/W$^{1/2}$
\end{itemize}


For ideal PMSMs, $K_T = K_B$ on the q-axis; our slight deviations likely resulted from imperfectly sinusoidal back-EMF profiles.

\subsection{Dynamic Properties}
\label{subsec:resultdynamic}

\begin{figure}
\begin{center}
    \includegraphics[width=.95\columnwidth]{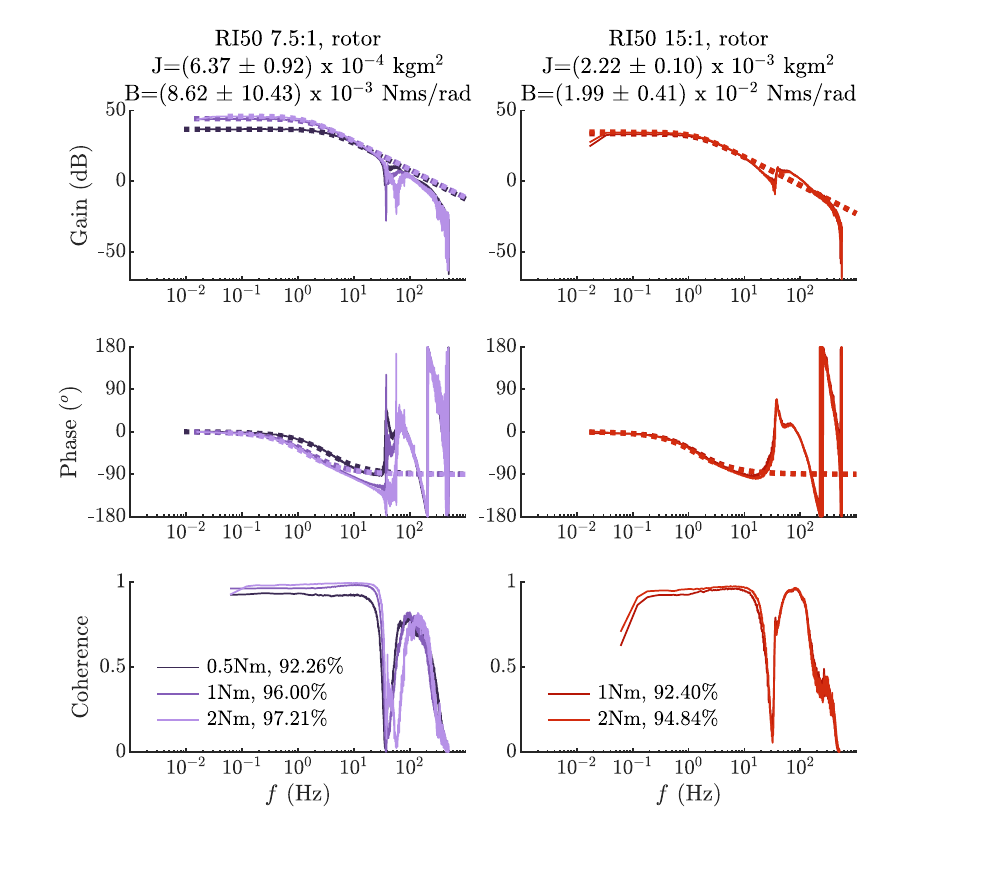}
    \caption{Mechanical System ID Frequency Response and Coherence. Legend indicated input amplitude and Variance Account For goodness of fit. The confidence intervals for the $J$ and $B$ are 2 standard deviations of identified $J$ and $B$ across input magnitudes}
    \label{fig:mech_sysid}
\end{center}
\end{figure}

The frequency domain non-parametric estimate and high coherence of $>0.9$ up to 40~Hz (Fig. \ref{fig:mech_sysid}) suggested that the torque and velocity data could be explained by a first-order linear model:
\begin{equation}
    \frac{\Omega (s)}{ T_b (s)} = \frac{1}{J_a s + B_a}
    \label{eq:mech_dyn_model}
\end{equation}
where $\Omega(s)$ is actuator angular velocity, $T_b (s)$ is backdriving torque, $J_a$ is actuator output inertia [kgm$^2$], $B_a$ is actuator viscous damping [Nms/rad], and $s$ is the Laplace variable.
The following parameters resulted from fitting the model: the actuator inertia $J_a$ was estimated as $(6.37 \pm 0.92) \times 10^{-4}$ kgm$^{2}$ for the 7.5:1 transmission and $(2.22 \pm 0.10) \times 10^{-3} $ kgm$^2$ for the 15:1 transmission.
The goodness of fit is evaluated with the Variance Account For (VAF) metric.
The parametric fit showed $92\%-98\%$ VAF, indicating a high-quality parameter estimation, and the estimates lined up very closely to the predictions from the equation for actuator inertia, $J_a$:
\begin{equation}
    J_a \doteq N^2 J_m
    \label{eq:act_Jm}
\end{equation}
where the transmission ratio, $N$, represents a speed reduction and $J_m$ represents motor inertia (found in \cite{Urs2022Framework}).

The drop in the model fit quality above 40Hz (Fig. \ref{fig:mech_sysid}) coincides with the cutoff frequency of the low-pass filter applied to the system input (\ref{subsec:methoddynamic}).
The model overall sufficiently explains the actuator mechanical dynamics, especially for application in legged robots, in which leg and torso dynamics are typically slower than 40Hz.

\subsection{Thermal Modeling}
\label{subsec:resultthermal}

\subsubsection{Electrical-Thermal Calibration and Measurement}
\label{subsubsec:resultthermistor}
Winding temperature $\overline{T_w}$ [$\degree$C] and resistance $R_{ll}$ varied linearly: 
\begin{equation}
    \overline{T_w} = 139.564R_{ll} - 177.826
\end{equation}
with $R^2 = 0.99$.

This relationship can also be stated as a 0.51\% increase in resistance per \degree C, which is similar to common values for pure copper (the winding material) of 0.43\%/\degree C \cite{noauthor_resistivity_nodate}. 
Discrepancies are due to incorporating the non-heated winding temperature into the temperature average, and potentially the enamel coating on the windings and non-copper elements in the composition.

\begin{figure}
\begin{center}
\includegraphics[width=.9\columnwidth]{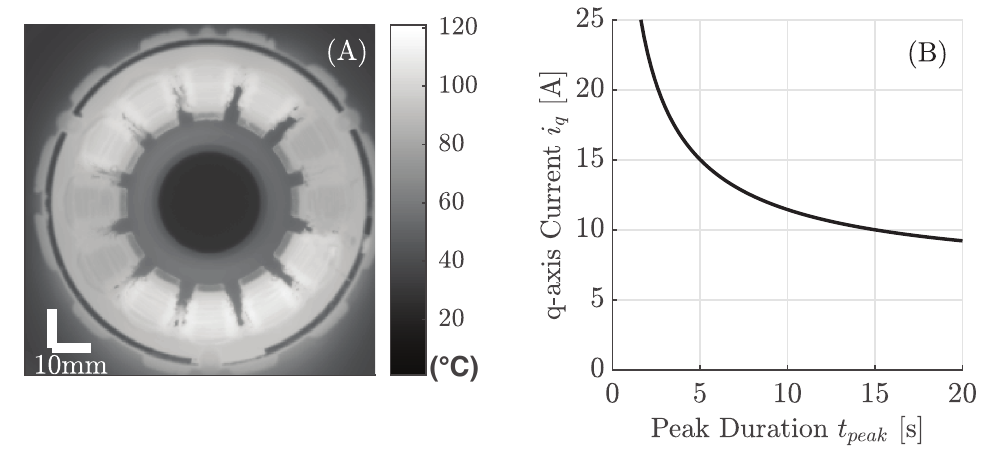}
    \caption{\textbf{(A)} Averaged thermal image over 50 frames of motor during electrical-thermal calibration. Eight of the twelve windings appear hotter than the rest, indicating a Wye winding style. \textbf{(B)} A plot showing the peak duration any current can be sustained predicted using the thermal model. A 9.2 A peak may be sustained for 20s or a 22 A peak may be sustained for 2s without exceeding temperature limits.}
    \label{fig:thermal_bw}
\end{center}
\end{figure}

\subsubsection{Thermal Dynamics}
\label{subsubsec:resultthermdyn}

Random-input thermal testing revealed that the cooling fans provide an approximately 3.5$\times$ cooling advantage, bringing performance limits to near parity with metallic actuators \cite{minicheetah, lee_empirical}.
A transfer function model (\ref{eq:thermal_tf}) fit to the thermal data yielded the system parameters in Table \ref{tab:therm_terms}; the total thermal resistance $R_{th}$ was 2.27 $^\circ$C/W.

\begin{table}
\begin{center}
\caption{Thermal Model Parameters}
    \begin{tabular}{ c c | c c } 
        & & Fan & No Fan  \\
        \hline \hline
        $R_{WH}$ & [\degree C/W]  & 0.828 & 1.09  \\
        $R_{HA}$ & [\degree C/W]  & {1.44} & {6.68}  \\
        $C_{WH}$ & [J/\degree C]  & 15.9 & 18.4 \\
        $C_{WA}$ & [J/\degree C]  & 146 & 138 \\
        \label{tab:therm_terms}
    \end{tabular}
    
\end{center}

\end{table}

The 3.5$\times$  reduction in $R_{th}$ with the fans nearly doubles the amount of continuous current that can be safely dissipated, corresponding to a similar near doubling in actuator torque capabilities.
The identification allows us to establish limits (Fig. \ref{fig:thermal_bw}B) and shows that the thermal capacitances with the fans on or off are very similar, as expected.



\subsection{Power Efficiency}
\label{subsec:resultpower}

\begin{figure}
\begin{center}
    \includegraphics[trim=0.2cm 0.0cm 0.1cm 0.0cm, clip,width=.95\columnwidth]{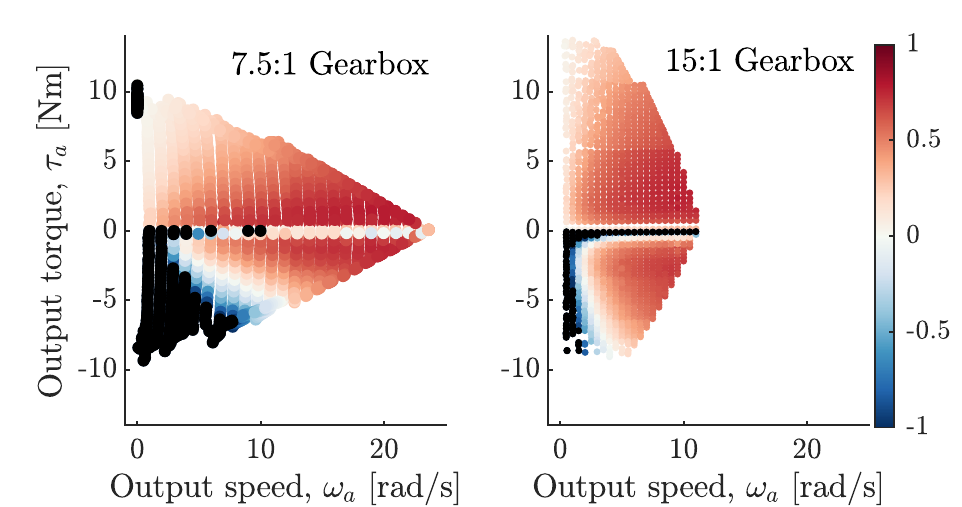}
    \caption{Efficiency map of the two actuator designs. 
    In slower and higher-torque regimes, the 15:1 transmission is more efficient. 
    In the positive work quadrant ($\tau_a>0$), efficiency is the mechanical output divided by the electric input power.
    In the negative work quadrant, the definition is inverted.
    Negative efficiencies imply actuator sinks power from both the electrical and mechanical sides. 
    Efficiencies $<-1$ are black. }
    \label{fig:power}
\end{center}
\end{figure}

Power efficiency testing yielded trends that match up well with the literature for other actuators \cite{lee_empirical}: efficiency is high at high speeds and low torques, but efficiency is reduced at low speed and high torque regimes by joule heating losses, which scale with $i_q^2$ and therefore $\tau_m^2$ (Fig. \ref{fig:power}).

\subsection{Lifecycle}
\label{subsec:resultlifecycle}

Lifecycle testing yielded highly positive results: in total, two 7.5:1 actuators were tested with gait-like test conditions for 57 hours with no failures and only limited performance degradation.
This length of time corresponds to $\sim$420,000 strides of Mini-Cheetah in the source data. 

\begin{figure}
\begin{center}
\includegraphics[trim=0.2cm 0.0cm 0.2cm 0.0cm, clip, width=.9\columnwidth]{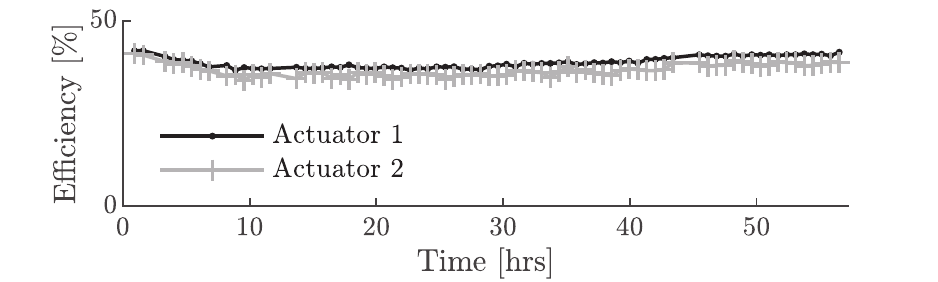}
    \caption{Average positive work efficiencies during the 57 hour lifecycle test of two 7.5:1 actuators. 
    An initial drop in efficiency of both actuators occurs over 8-10 hours before a leveling off, followed by a rebound to near original levels.}
    \label{fig:lifecycle_eff}
\end{center}
\end{figure}

The power efficiency interval tests (see Section \ref{subsec:methodpower}) revealed a clear trend: positive work efficiency steadily dropped for the first 8--10 hours of testing before leveling-off at a 3.5\% lower value (Fig. \ref{fig:lifecycle_eff}).
As testing went on, fine plastic dust built up in the transmission, potentially increasing friction up to a saturation point, causing a drop-then-level trend.
The efficiency then rebounded to near-original values, suggesting the presence of a slower process, such as breaking-in of parts or bearings.

The damping interval tests showed approximately constant damping coefficients: for a control input of 2.5~rad/s, the damping coefficient was $76\pm 12$~mNms/rad and for 5.0 rad/s, it was $41\pm 8$~mNms/rad without a clear trend.

For the gait data playback and the interval tests, encoder data from both sides of the dynamometer were collected.
Assuming no encoders drift or errors, differences between the encoder readings must be due to torsional flexion and backlash.
We observed a maximum of 56~mrad difference in the first hour, which grew to 82~mrad in the last hour.
Assuming that the Young’s modulus of the components did not change during the test, this 26~mrad growth in discrepancy must be entirely due to an increase in backlash (average of 13~mrad per actuator).
The rate of backlash growth was approximately linear at a rate of $250\pm 30$~{$\mu$}rad/hr per actuator.

\subsection{Control Performance}
\label{subsec:resultcontrol}

\begin{figure}
\begin{center}
\includegraphics[width=.9\columnwidth]{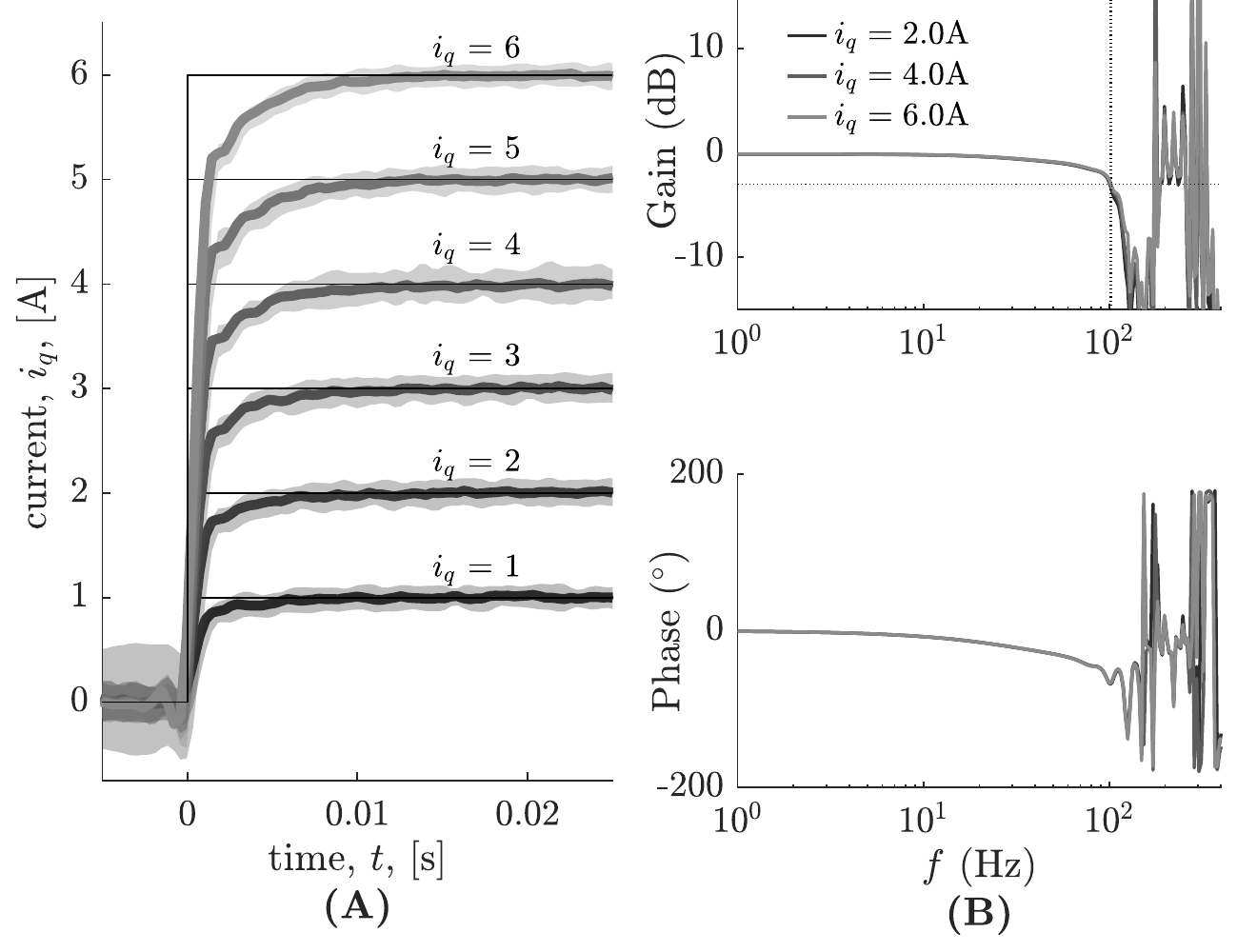}
    \caption{(A) Step response of q-axis current controller with an average rise time of 2.65ms (B) Frequency domain current control response at 3 input amplitudes, with with a half-power bandwidth $f_c$ of 102~Hz.}
    \label{fig:step_response}
\end{center}
\end{figure}

Step-response tests of the motor driver's built-in manually-tuned current controller yielded a rise time of $2.65\pm 1.88$~ms (Fig. \ref{fig:step_response}A)
The random-input frequency response indicated a half-power bandwidth of $102\pm1$~Hz for the current controller (Fig. \ref{fig:step_response}B) and $11-30$ Hz for the position controller (plot omitted).
The current controller bandwidth is lower than comparable actuators, likely due to the high motor inductance $L_e$. 

\section{Discussion}
\label{sec:discussion}

We present a full characterization of the mechanical, thermal and electrical properties of two open-source, fully 3D printed, quasi-direct drive actuators. 
With our thermal solution, actuator mechanical torque capabilities were comparable to metallic actuators currently used for small quadrupedal robots
, but with a significantly lower inertia \cite{minicheetah, lee_empirical}.
Life-cycle testing of the 7.5:1 planetary transmission prototype for 420,000 strides (57~hrs) revealed little increase in wear and backlash---after a minor drop in the first 10~hrs, efficiency rebounded to original levels.
This pattern suggests a relatively quick rise in friction until saturation, until a slower break-in process became apparent; nylon gears exhibit a similar process \cite{zhang_physical_2020}. 

Our 3D printed actuator designs reduce time and cost barriers, thereby contributing to the growing field of open-source robot development \cite{azocar2020design, grimminger_open_2020}. 
These two designs exemplify the reliability, high performance, and durability of 3D printed actuators at multiple scales.
Bio-inspired designers can extend this work to create actuators and robots that mimic a broad range of animal capabilities \cite{Urs2022SICB} ,  
while being confident that they can withstand rigorous and sustained use in research settings.
We plan to develop future versions with transmission ratios optimized for particular legged robot tasks and 3D printed series-elastic actuators. 

\section*{ACKNOWLEDGMENT}
The authors thank C Bradley (ME Machine Shop) for fabrication assistance; J Pieper (mjbots) for motor driver support; and M Ghaffari (CURLY Lab) for providing Mini-Cheetah data.

Designs are available on \url{www.embirlab.com/3dpactuator}.

\bibliographystyle{IEEEtran}
\bibliography{root.bib}

\end{document}